\documentclass[lettersize,journal]{IEEEtran}
\usepackage{amsmath,amsfonts}
\usepackage{algorithmic}
\usepackage{algorithm}
\usepackage{array}
\usepackage[caption=false,font=normalsize,labelfont=sf,textfont=sf]{subfig}
\usepackage{textcomp}
\usepackage{stfloats}
\usepackage{url}
\usepackage{verbatim}
\usepackage{graphicx}
\usepackage{cite}

\usepackage{times}
\usepackage{latexsym}
\usepackage{amssymb}
\usepackage{amsthm}
\usepackage{booktabs}
\usepackage{multirow}
\usepackage{soul}
\usepackage{bbding}

\usepackage{color,xcolor}

\usepackage{makecell}
\newcommand{\tabincell}[2]{\begin{tabular}{@{}#1@{}}#2\end{tabular}}

\usepackage[T1]{fontenc}

\usepackage{adjustbox}
\usepackage{tabularx}
\usepackage{float}
\usepackage{pythonhighlight}

\newcommand{\ourmethod}{\textsc{LC-ICL}}

\begin{document}

\title{\ourmethod{}: Label-Guided Contrastive In-Context Learning for Robust Information Extraction}

\author{Xiao You$^{1}$, Tianwei Yan$^{2}$, Shan Zhao$^{1}$\thanks{Corresponding author: Shan Zhao.}\\
$^{1}$Hefei University of Technology, Hefei, China\\
$^{2}$Chongqing Jiaotong University, Chongqing, China}

\maketitle

\begin{abstract}

There has been increasing interest in exploring the capabilities of advanced large language models (LLMs) in the field of information extraction (IE), specifically focusing on tasks related to named entity recognition (NER) and relation extraction (RE).
Although researchers are exploring the use of few-shot information extraction through in-context learning with LLMs, they tend to focus only on using correct or positive examples for demonstration, neglecting the potential value of incorporating incorrect or negative examples into the learning process.
In this paper, we present \ourmethod{}, a novel few-shot technique that leverages both correct and incorrect sample constructions to create in-context learning demonstrations. This approach enhances the ability of LLMs to extract entities and relations by combining positive samples with negative samples annotated by error-cause labels. These labels expose more detailed error features in erroneous examples, enabling the model to understand why similar predictions fail and avoid repeating such errors during inference.
Specifically, our proposed method taps into the inherent contextual information and valuable information in hard negative samples and the nearest positive neighbors to the test and then applies the in-context learning demonstrations based on LLMs. 
Our experiments on various datasets indicate that \ourmethod{} outperforms previous few-shot in-context learning methods, delivering substantial enhancements in performance across a broad spectrum of related tasks. 
These improvements are noteworthy, showcasing the versatility of our approach in diverse scenarios.
\end{abstract}

\section{Introduction}
Information extraction (IE) is an important task in natural language processing, aiming to obtain structured knowledge from plain text. It can be applied across different domains, such as knowledge graph construction \cite{KG-survey-2023} and question answering systems \cite{QA-2006}. With the rise of large language models (LLMs) \cite{gpt3-2020,ICL-2022,llama-2,gpt4-2023}, IE has achieved remarkable progress \cite{gpt-ie-li-2023,llm-ie-survey-2023}. Recent advances in few-shot IE have shifted the focus from traditional supervised fine-tuning methods to leveraging LLMs for in-context learning (ICL) demonstrations \cite{ICL-NER-2023,Z-ICL-2023}.
\begin{figure}[t]
\begin{center}
    \includegraphics[width=1\columnwidth]{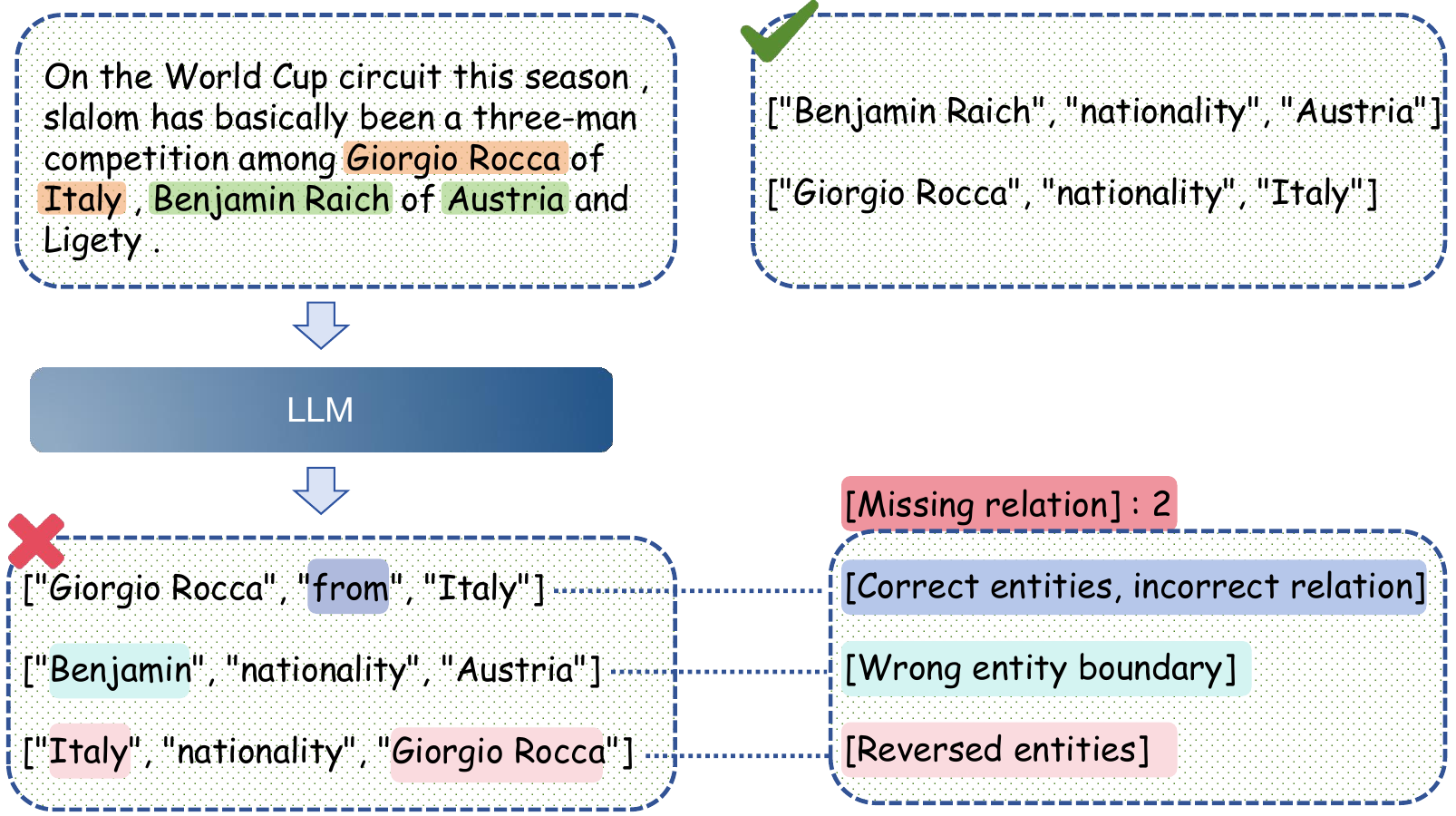}
    \vspace{-2mm}
    \caption{Using the relation extraction (RE) task as an example, this figure illustrates the core mechanism of \ourmethod{}. When performing direct inference, large models are prone to generating a large number of errors with specific patterns. \ourmethod{} introduces an error labeling system that feeds back error types to the model, guiding it to refine its generation, thereby significantly reducing errors and improving inference accuracy.} 
    \label{Fig.intro} 
    \vspace{-6mm}
\end{center}
\end{figure}
Previous work \cite{ChatIE-2023, ICL-NER-2023, CoT-ER-2023, Fine-Tuned+Flan-T5+CoT} has explored using natural language prompts or in-context learning (ICL) demonstrations to guide LLMs in annotating test data under few-shot settings, sometimes requiring additional pretraining or fine-tuning steps. To better align with the structured nature of information extraction tasks, more recent approaches \cite{CodeIE-2023, GoLLIE-2023, InstructUIE-2023, Code4Struct-2023, GPT-RE-23} employ code-like or structured prompts to enhance the consistency between pretraining and inference. However, these methods have yet to fully unlock the potential of LLMs, partly because the models rely on limited positive data and cannot learn from their own errors.

To address this issue, this paper proposes a contrastive in-context learning approach that leverages both positive and negative examples to extend the learning process of LLMs, thereby exposing them to a wider range of scenarios, including typical errors. The method is designed to exploit the value of often overlooked negative data, enabling more comprehensive and robust information extraction capabilities.

Assume that the model has already learned task execution and problem-solving patterns from the positive information extraction dataset, yet its predictions still contain errors. In this case, the model should reflect on the causes of these errors, categorize the error types, and attempt to avoid them in subsequent reasoning. Therefore, incorporating information related to negative samples can help address this issue. Inspired by this idea, this paper integrates both correct/positive and incorrect/negative examples into ICL demonstrations to enhance the information extraction performance of in-context learning.

Specifically, we first employ a large-scale model to generate labels for the annotated data in order to select hard negative samples. Then, we select semantically similar positive samples from the training data for the current test instance, and design the most suitable in-context demonstrations using different models (natural language LLMs or code LLMs). In the module for selecting error/negative samples that contain richer knowledge, we adopt a semantic-similarity-aware retrieval approach for ranking.

To demonstrate the advantages of the proposed method, we conduct experiments on three named entity recognition (NER) and five relation extraction (RE) benchmark datasets, followed by a thorough analysis of the benefits brought by this approach.

The main contributions of this paper are summarized as follows:
\begin{itemize}
\item We propose \ourmethod{}, a contrastive in-context learning approach that incorporates positive demonstrations and negative samples annotated with error-cause labels. These labeled negative samples provide more detailed error features, helping LLMs recognize failure patterns and avoid similar mistakes during information extraction.
\item We design an effective retrieval strategy to select hard negative samples as part of in-context learning, leveraging them to further strengthen information extraction capabilities.
\item We conduct extensive experiments on benchmark datasets, demonstrating the effectiveness and broad applicability of the proposed method across NER and RE tasks.
\end{itemize}

\section{Task Formulation}

Given a sentence $X$ containing $l$ tokens $x_1, x_2, \cdots, x_l$, the goal of IE tasks is to predict a structured output $Y$ (named entities or relations) from $x$. In the NER task, the target $Y$ is entity spans with entity types $E((e,t)|x_i,\dots,x_j)$, where $e$ is an entity in the sequence, and $t$ is an entity type from a predefined set of entity types $\mathcal{T}$ (e.g., \texttt{LOC}, \texttt{PER}, \texttt{ORG}).

In the RE task, the target $Y$ is a set of relations between entities, typically expressed as triples $(e_1, r, e_2)$. This not only involves predicting the relation $r \in \mathcal{R}$ but also includes the types $t_1$ and $t_2$ of entities $e_1$ and $e_2$, where $\mathcal{R}$ represents relation types (e.g., \texttt{Work For}, \texttt{Live In}, \texttt{Located In}). The types of entities $e_1$ and $e_2$ also need to be predicted, where $t \in \mathcal{T}$ represents entity types.

We formulate IE as a generation task, prompting large language models to perform inference for NER or RE.

In the few-shot in-context learning setting for information extraction, the demonstrations are structured into four components: (1) negative examples, (2) positive examples, (3) instructions, and (4) test text. The output of LLMs is a list of tuples, which in the NER task is $[(e_1, t_1), \dots,(e_j,t_j)]$, such as [('Steve', 'person')].

\begin{figure*}[!t]
\begin{center}
    \includegraphics[width=0.95\linewidth]{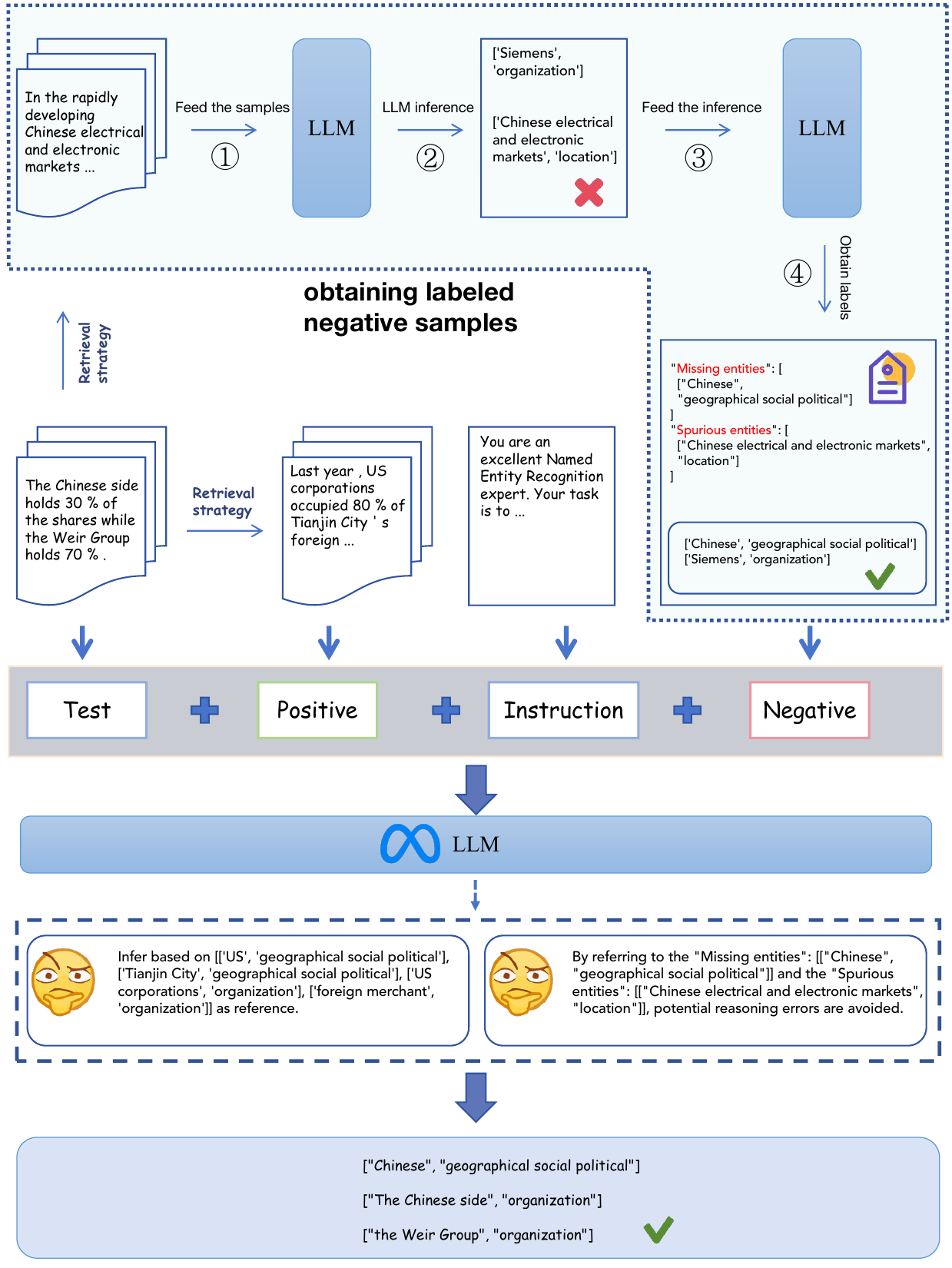}
    \vspace{-1mm}
    \caption{The figure illustrates an overview of the \ourmethod{} framework for information extraction tasks, using the Named Entity Recognition (NER) task as an example. The method constructs positive and negative examples, which, along with the test samples and task instructions, are input into the large language model to generate predictions. Steps 1--4 represent the process of obtaining labeled negative samples based on retrieved examples.}  
    \label{Fig.frame_main} 
    \vspace{-5mm}
\end{center}
\end{figure*}

\section{\ourmethod{}}
\subsection{Model Overview}

As shown in Figure \ref{Fig.frame_main}, in the context examples, our method mainly consists of 4 parts: Instruction part, Test sentence part, Positive Samples part, and Negative Samples part.

Unlike prior methods that rely exclusively on positive samples for in-context learning, our approach, \ourmethod{}, leverages both positive examples and negative instances annotated with explicit reasoning errors. For the selection of positive samples, we employ different retrieval strategies to select samples from the training set that correspond to the test examples. For negative sample selection, we first sample a subset of data from the training dataset of each dataset to form a negative sample collection. Then, $M_{\mathrm{infer}}$ generates predictions for the data in the negative sample collection, and $M_{\mathrm{label}}$ compares each prediction with its gold label to assign error labels, forming the final negative sample pool. Finally, we also use different retrieval strategies to retrieve a negative sample from the negative sample pool that corresponds to the test sample. Combined with the previously selected positive samples, we use both positive and negative samples together as examples for ICL to the large model. The large model will learn correct patterns from positive examples, thereby improving accuracy, while negative samples with error labels will inform the model about similar error patterns, helping it avoid similar mistakes and reduce error rates.

\subsection{LC-ICL Demonstrations Construction}
We construct a prompt for each given test sentence, and input it into the LLM. Each prompt contains the following components:

\textbf{Test Sentence}
${X_t}$ represents the original text of the RE/NER test sample, which is used for the final evaluation of the large model's performance in executing RE/NER tasks under the specified method.

\textbf{Instruction}
$\mathcal{I}$ We constrain the large model to different expert roles according to different tasks, and based on the task differences, provide the relation set $\mathcal{R}$ or entity type set $\mathcal{T}$. We use brief commands to require the large model to complete RE/NER tasks, and need it to return responses in a fixed list format. The instruction further guides the model to attend to negative samples with error-cause labels, so that it can acquire fine-grained error-feature signals from erroneous examples and avoid making similar mistakes on the test sentence.

\textbf{Positive Samples}
$\mathcal{P}$ Referring to previous work by \cite{CodeIE-2023}, we use the KNN algorithm (k-nearest neighbors algorithm) to select examples from the training set that have the highest semantic similarity with the test sample $X_{t}$ in the embedding space, to serve as positive examples $\mathcal{P}$.

\textbf{Negative Samples}
$\mathcal{N} $
Let $M_{\mathrm{infer}}$ denote the inference model that produces predictions, and let $M_{\mathrm{label}}$ denote the label-construction model that assigns error labels by comparing each prediction with its gold label.

\begin{equation}
\mathcal{X}'_{\mathrm{pred}} = M_{\mathrm{infer}}(\mathcal{I}, \mathcal{X}'_{t})
\label{eq:xi}
\end{equation}
\begin{equation}
\mathcal{N}_{\mathrm{label}} = M_{\mathrm{label}}(\mathcal{I}, E, \mathcal{X}'_{t}, \mathcal{X}'_{\mathrm{pred}}, \mathcal{X}'_{g})
\label{eq:nlabel}
\end{equation}
\begin{equation}
\mathcal{N} = \{(\mathcal{X}'_{t}, \mathcal{X}'_{\mathrm{pred}}, \mathcal{X}'_{g}, \mathcal{N}_{\mathrm{label}})\}
\label{eq:n}
\end{equation}
Following the approach in previous work \cite{CodeIE-2023}, we first sample 2500 labeled examples from the training set and use $M_{\mathrm{infer}}$ to generate predictions for these samples, as shown in Equation \ref{eq:xi}. Here, $\mathcal{X}'_{t}$ represents the sampled input examples, $\mathcal{X}'_{g}$ represents their corresponding gold labels, and $\mathcal{X}'_{\mathrm{pred}}$ represents the predictions produced by $M_{\mathrm{infer}}$. Next, we provide the input examples, predictions, and gold labels to $M_{\mathrm{label}}$ to annotate incorrect predictions with error labels, as shown in Equation \ref{eq:nlabel}. Here, $E$ represents the set of error labels, and $\mathcal{N}_{\mathrm{label}}$ represents the generated error label set. Finally, using the same method used to filter positive samples, we employ the KNN algorithm to select negative samples corresponding to the test samples, and store $\mathcal{X}'_{t}$, $\mathcal{X}'_{\mathrm{pred}}$, $\mathcal{X}'_{g}$, and $\mathcal{N}_{\mathrm{label}}$ together in the negative sample pool, as shown in Equation \ref{eq:n}.

\begin{equation}
Y_{t} = M_{\mathrm{infer}}(\mathcal{P}, \mathcal{N}, \mathcal{I}, X_{t})
\label{eq:yt}
\end{equation}
Overall, the input and output of our proposed framework can be represented by formula \ref{eq:yt}, where $Y_{t}$ represents the inference result produced by $M_{\mathrm{infer}}$ for the test sample $X_{t}$.

\begin{equation}
E_{\mathrm{RE}} = \{E_1, E_2, E_3, E_4, E_5, E_6\}
\label{eq:ere}
\end{equation}
\begin{equation}
E_{\mathrm{NER}} = \{E_1', E_2', E_3', E_4', E_5'\}
\label{eq:ener}
\end{equation}
\begin{equation}
E_{\mathrm{RE}} \cup E_{\mathrm{NER}} \subseteq E
\label{eq:union}
\end{equation}

We carefully analyze the causes of reasoning errors and formulate detailed error labels for RE and NER tasks respectively. Errors in RE are categorized into 6 types, as shown in Equation \ref{eq:ere}, while errors in NER are divided into 5 types, as shown in Equation \ref{eq:ener}. In Equation \ref{eq:union}, $E_{\mathrm{RE}}$ and $E_{\mathrm{NER}}$ together constitute $E$.

The specific label types and their meanings in Equations \ref{eq:ere} and \ref{eq:ener} can be found in Appendix~\ref{sec:appendix_b}.

\subsection{LC-ICL Retrieval Strategy}
\subsubsection{KNN-Based Retrieval}

In few-shot learning, selecting demonstrations that are semantically similar to the test sample is crucial \cite{KATE-knn-2022}. Research has shown that using KNN methods to retrieve the most similar examples from the training set can improve model performance \cite{ICL-Think-2022,KATE-knn-2022,Code4UIE-2023}. KNN relies on effective embedding spaces to encode inputs and examples; therefore, pre-trained language models or improved sentence embedding methods have been proposed to optimize sentence representations.

For information extraction tasks, we also adopt a retrieval method based on sentence embeddings, utilizing the k-nearest neighbor algorithm to filter out demonstration sentences from the training set that are relevant to the target task. By calculating semantic similarity, we select the top k sentences containing entities or relations, and use LLMs combined with cosine similarity for matching, thereby providing higher quality demonstrations for in-context learning.

\subsubsection{Error Label Retrieval}

\begin{algorithm}[!t]
    \renewcommand{\algorithmicrequire}{\textbf{Input:}}
    \renewcommand{\algorithmicensure}{\textbf{Output:}}
    \caption{Negative Sample Generation and Retrieval Algorithm}
    \begin{algorithmic}[1]
            \REQUIRE Test sample $X_t$; labeled training dataset $\mathcal{D}_{\text{train}}$; inference model $M_{\mathrm{infer}}$; label-construction model $M_{\mathrm{label}}$; error label set $E$; sample size $k=2500$.
            
            \ENSURE Retrieved negative sample $\mathcal{N}$ for test sample $X_t$
            
            \STATE /* Phase 1: Generate negative sample pool */
            \STATE $\mathcal{X}' \leftarrow \text{Sample}(\mathcal{D}_{\text{train}}, k)$ \COMMENT{Sample $k$ labeled instances}
            \STATE Initialize negative sample pool $\mathcal{S} \leftarrow \emptyset$
            
            \FOR{each labeled sample $(x'_t, x'_g) \in \mathcal{X}'$}
                \STATE $x'_{\mathrm{pred}} \leftarrow M_{\mathrm{infer}}(\mathcal{I}, x'_t)$ \COMMENT{Generate prediction as in Eq.~\ref{eq:xi}}
                
                \STATE $\mathcal{N}_{\text{label}} \leftarrow M_{\mathrm{label}}(\mathcal{I}, E, x'_t, x'_{\mathrm{pred}}, x'_g)$ \COMMENT{Generate error labels using the gold label}
                \STATE $\mathcal{S} \leftarrow \mathcal{S} \cup \{(x'_t, x'_{\mathrm{pred}}, x'_g, \mathcal{N}_{\text{label}})\}$

            \ENDFOR
            
            \STATE /* Phase 2: Retrieve relevant negative sample for test instance */
            \STATE Compute semantic embedding for $X_t$ and all samples in $\mathcal{S}$
            \STATE $\mathcal{N} \leftarrow \text{KNNRetrieval}(X_t, \mathcal{S})$ \COMMENT{Retrieve most similar negative sample}            
            \RETURN $\mathcal{N}$
    \end{algorithmic}
    \label{alg:1}
\end{algorithm}

Inspired by \cite{faghri2017vse++}, in addition to using positive examples to help large models recognize correct answers, negative examples also serve an instructive role, helping models correct similar errors. Mo et al. \cite{mo2024c-icl} identified hard negative examples by querying the model multiple times and selecting high-confidence erroneous outputs. However, repeatedly calling the large model to obtain hard negative samples consumes significant time and incurs expensive costs. Therefore, we propose an error label retrieval strategy. We conducted a detailed analysis of the reasons for large model inference errors on NER/RE datasets and formulated detailed error labels based on these reasons, as shown in Algorithm \ref{alg:1}. After a single forward pass on the sampled training data using $M_{\mathrm{infer}}$, we provide each input, its prediction, and its gold label to $M_{\mathrm{label}}$ to obtain error labels. The resulting negative sample pool stores the input, prediction, gold label, and error labels together. Finally, we use the kNN algorithm to select semantically similar labeled error samples as negative examples for test samples.

\section{Experiments}
\subsection{Datasets}
\noindent\textbf{RE Datasets}
For relation extraction, we evaluate on datasets CoNLL04 \cite{conll2004}, NYT10\cite{riedel2010modeling},NYT11 \cite{takanobu2019hierarchical} ,SciERC \cite{scierc} and ADE\cite{gurulingappa2012development}. 
We adopt the dataset splits from prior UIE work~\cite{UIE-2022} for all these datasets. 

\noindent\textbf{NER Datasets}
We evaluate our approach on NER task with ACE04\cite{ace04} , ACE05\cite{ace05} and NCBI\cite{dougan2014ncbi}. and we split the datasets followed by the works \cite{MRC-NER-2020,moying-2023,mcl-ner-moying-2023,CodeIE-2023}.
Table \ref{tab:datasets_anlasis} shows the dataset statistics in Appendix \ref{sec:appendix_a}.

\subsection{Experiments Setting}

We use Llama \cite{grattafiori2024llama} as the backbone for $M_{\mathrm{infer}}$ on test sets and DeepSeek \cite{liu2024deepseek} as the backbone for $M_{\mathrm{label}}$ when constructing error labels for negative samples.

For the IE task, we construct both positive and negative samples to serve as contextual prompts. For positive sample selection, we employ a KNN retrieval strategy to select the most matching positive sample from the training set for each test sample. For negative samples, we first perform inference on the sampled training instances using $M_{\mathrm{infer}}$, then present each input, its prediction, and its gold label to $M_{\mathrm{label}}$ to obtain error labels, forming a negative sample pool that stores the gold label together with the prediction and error labels. Finally, using the same KNN retrieval strategy, we retrieve the most similar sample from the negative sample pool as the negative sample for each test case. Due to the large number of samples in the training datasets, constructing error labels for all training data would be costly. Following \cite{CodeIE-2023}, we randomly sample 2,500 examples from each training dataset to build the negative sample pool for subsequent negative sample retrieval.

Detailed experimental settings are provided in Appendix \ref{sec:appendix_a}.

\subsection{Evaluation}

Following previous work by \cite{UIE-2022, CodeIE-2023}, we employ a strict evaluation metric for both NER and RE F1 scores. In the NER datasets, a prediction is considered correct only when both the entity name and type are correct. Similarly, in the RE datasets, a prediction is counted as correct only when both the entities and their relationship are accurately identified. Additionally, we introduce an instruction-following metric, which allows us to observe whether errors in model reasoning stem from poor instruction adherence or from difficulties in the task execution itself. To ensure the reliability of our results, we conducted three rounds of experiments with three different random seeds and reported the average scores.

\subsection{Results}
We investigate the performance of our proposed method in comparison to traditional in-context learning  approaches and direct inference on NER and RE tasks.
In addition, we conduct experiments with two model scales, Llama-3.1-8B-Instruct and Llama-3.3-70B-Instruct, to further examine the impact of model size on performance.
\begin{table*}[ht]
\caption{Performance comparison on RE benchmarks in terms of F1 score (\%). ICL denotes the model performing in-context learning without parameter updates. PE refers to using only positive examples, which is the traditional ICL, while LC-ICL, our method, includes both positive and negative examples.}
\label{tab:main_results_RE}
\begin{adjustbox}{width=0.92\linewidth,center}
\begin{tabular}{lcccccccc}
\toprule
\multicolumn{1}{c}{\multirow{2}{*}{Model}} & \multirow{2}{*}{Paradigm} & \multirow{2}{*}{Backbone} & \multicolumn{5}{c}{RE} \\ 
\multicolumn{1}{c}{}                       &                           &                           & CoNLL04                              & NYT10     &NYT11                  & SciERC   &ADE            \\ \hline
DI              & w/o ICL                      & Llama-3.1-8B-Instruct         & 0.35               & 0.27                & 0.09              & 0.00\footnote{Repeated verification confirms that the model consistently returns empty results on this dataset, resulting in an F1 score of 0\%.}          & 0.79        \\ 
PE              & ICL                      & Llama-3.1-8B-Instruct         & 15.11                & 15.73                & 8.28               & 2.02            & 13.60        \\ 
\textbf{LC-ICL (ours)}    & ICL                      & Llama-3.1-8B-Instruct             &\textbf{32.27}                &\textbf{38.69 }              &\textbf{21.52}                & \textbf{4.18}          &\textbf{18.10}       \\
\hline

DI                   & w/o ICL                    & Llama-3.3-70B-Instruct            & 7.02               & 6.33                & 6.61                & 2.78  & 0.68              \\

PE                   & ICL                        & Llama-3.3-70B-Instruct            & 48.42               & 27.77                & 23.01                & 9.35  & 49.62               \\
\textbf{LC-ICL (ours)}               & ICL                      & Llama-3.3-70B-Instruct         & \textbf{50.71}                & \textbf{42.04}                & \textbf{34.15}               & \textbf{12.21}            &\textbf{54.13}         \\ 
                                        
                                        \bottomrule
\end{tabular}
\end{adjustbox}

\end{table*}
\begin{table*}[ht]
\caption{Performance comparison on NER benchmarks in terms of F1 score (\%). w/o ICL represents zero-shot performance, while ICL denotes in-context learning. DI refers to direct inference, contrasting with PE, which uses only positive examples as in traditional ICL. Our method, LC-ICL, incorporates both positive and negative examples.}
\label{tab:main_results_NER}
\begin{adjustbox}{width=0.92\linewidth,center}
\begin{tabular}{lcccccl}
\toprule
\multicolumn{1}{c}{\multirow{2}{*}{Model}} & \multirow{2}{*}{Paradigm} & \multirow{2}{*}{Backbone} & \multicolumn{3}{c}{NER}   &  \\ 
\multicolumn{1}{c}{}                       &                           &                           & ACE04 & ACE05 &NCBI&  \\ \hline

DI             & w/o ICL                       & Llama-3.1-8B-Instruct          &11.01   &16.88  &\textbf{39.40}   &  \\ 
PE             & ICL                       & Llama-3.1-8B-Instruct          &14.72    &24.59  & 33.30   &  \\ 

\textbf{LC-ICL (ours)}     & ICL                       & Llama-3.1-8B-Instruct             &\textbf{21.36}    &\textbf{25.76}  &\underline{34.36}    &  \\

\hline

DI                      &w/o ICL                       & Llama-3.3-70B-Instruct            & 24.66   &24.07  &40.44     &  \\

PE                      & ICL                       & Llama-3.3-70B-Instruct            & 37.18   &38.99  &38.20     &  \\

\textbf{LC-ICL (ours)}             & ICL                       & Llama-3.3-70B-Instruct            & \textbf{44.53}  &\textbf{44.64}  &\textbf{56.36}     &  \\

                                           \bottomrule
\end{tabular}
\end{adjustbox}
\end{table*}

\textbf{RE Results} Table \ref{tab:main_results_RE} shows the results of the RE task.

Overall, our method and the traditional method that only uses positive examples as ICL prompts (PE) both significantly outperform the direct inference method (DI), demonstrating that effective contextual examples can significantly enhance the model's ability to complete RE tasks. On the same Llama-3.1-8B-Instruct base model, our method outperforms the PE method because our approach not only uses positive examples to guide the large model in identifying similar patterns but also informs the model about potential errors. Through error label information, it avoids the hallucination content that might be produced when merely imitating positive example patterns using only positive samples. Furthermore, as the model parameter size increases, our method remains effective. On the Llama-3.3-70B-Instruct model, our method shows improvement over the ICL method across all RE datasets, further proving that our approach is effective not only on small-parameter models but also on large-parameter models.

On the Llama-3.1-8B-Instruct model, we conducted a detailed comparison between the PE method and our proposed LC-ICL method. On the NYT10 dataset, the traditional ICL method achieved an F1 score of only 15.73\%, while our method directly improved it to \textbf{38.69\%}, representing a relative improvement of 146\%. On the NYT11 dataset, the LC-ICL method (F1=\textbf{21.52\%}) showed a 159.8\% improvement compared to the PE method (F1=8.28\%), demonstrating our method's significant advantage on the NYT series datasets. On the SciERC scientific literature relation extraction task, the PE method performed extremely poorly with an F1 score of 2.02\%, while our method improved it to \textbf{4.18\%}. Although the absolute value is not high, the improvement exceeds 100\%, indicating that our Label design remains effective even in extremely low-resource scenarios. On the ADE medical relation extraction task, the PE method achieved an F1 score of 13.60\%, while our method improved it to \textbf{18.10\%}, demonstrating stronger relation modeling capabilities in the medical domain. On the CoNLL04 task, our method achieved an F1 score of \textbf{32.27\%}, a 113.7\% improvement compared to PE (15.11\%), fully validating the applicability and effectiveness of the LC-ICL strategy on classic small-scale RE datasets.

On the Llama-3.3-70B-Instruct model, we also compared the PE method with our proposed LC-ICL method. On the NYT10 dataset, the PE method achieved an F1 score of 27.77\%, while our method directly improved to \textbf{42.04\%}, representing a relative improvement of 51.4\%. On the NYT11 dataset, the LC-ICL method (F1=\textbf{34.15\%}) showed a 48.4\% improvement compared to the PE method (F1=23.01\%), continuing to maintain significant advantages on the NYT series datasets. On the SciERC scientific literature relation extraction task, the PE method achieved an F1 score of 9.35\%, showing relatively weak performance, while the LC-ICL method improved to \textbf{12.21\%}, a relative improvement of 30.6\%, indicating that the LC-ICL design is also effective on specialized domain texts. On the ADE medical relation extraction task, the PE method achieved an F1 score of 49.62\%, while the LC-ICL method improved to \textbf{54.13\%}, a relative improvement of 9.1\%, demonstrating stronger relation modeling capabilities in the medical domain. On the CoNLL04 task, the LC-ICL method achieved an F1 score of \textbf{50.71\%}, showing a 4.7\% improvement compared to PE (48.42\%). Although the improvement margin is relatively small, it further verifies the generalization ability of the LC-ICL strategy on large-scale models. Overall, the experimental results on the 70B model fully demonstrate that even on larger-scale models, the LC-ICL method incorporating negative examples can still bring significant performance improvements.

\noindent \textbf{NER Results} Table \ref{tab:main_results_NER} shows the results of the NER task. 

Similarly, in the NER task, our method and the PE method mostly outperform direct inference approaches. Compared with PE, LC-ICL generally achieves better results, suggesting that incorporating incorrectly predicted entities and their error-type labels as negative samples can help large language models avoid similar entity recognition errors. However, the gains are not uniform across all baselines and datasets; for example, direct inference remains competitive on the NCBI dataset with Llama-3.1-8B-Instruct.

Similarly, on the Llama-3.1-8B-Instruct model, we conducted a detailed comparison between the PE method and our approach. In the ACE04 dataset, the PE method only achieved an F1 score of 14.72\%, while the LC-ICL method significantly improved to \textbf{21.36\%}, representing a 45.12\% increase. In the ACE05 dataset, the PE method scored 24.59\%, with LC-ICL improving to \textbf{25.76\%}. Although the improvement margin was limited, it still demonstrates the robustness of the LC-ICL strategy across different datasets. For the NCBI disease entity recognition task, the PE method reached 33.30\%, while LC-ICL improved further to \textbf{34.36\%}. This result indicates that LC-ICL improves over PE on this medical-domain task, although direct inference remains a strong baseline on NCBI.

In the Llama-3.3-70B-Instruct model, our LC-ICL method demonstrates significant performance advantages. The experimental results show that our proposed LC-ICL method achieves F1 scores of \textbf{44.53\%}, \textbf{44.64\%}, and \textbf{56.36\%} in the three datasets, comprehensively outperforming other methods. Compared to PE, LC-ICL improves by 7.35 percentage points (a relative increase of 19.77\%) on the first dataset, 5.65 percentage points (a relative increase of 14.49\%) on the second dataset, and an impressive 18.16 percentage points (a relative increase of 47.54\%) on the third dataset. By comparing performance across different datasets, we find that the LC-ICL method exhibits strong robustness and adaptability, consistently maintaining its advantage on data with varying characteristics. These results convincingly demonstrate that our method's design philosophy effectively overcomes the limitations of traditional ICL methods, providing a more efficient paradigm for applying large language models to entity recognition tasks. Particularly on large-scale models like the 70B model, the LC-ICL method shows even more pronounced performance gains, indicating its ability to better leverage potential as model scale increases.

\section{Further Analysis}
\label{analysis}

\subsection{Ablation Study}

\begin{figure}[t]
\begin{center}
    \subfloat[RE Task]{
        \centering
        \includegraphics[width=0.95\columnwidth]{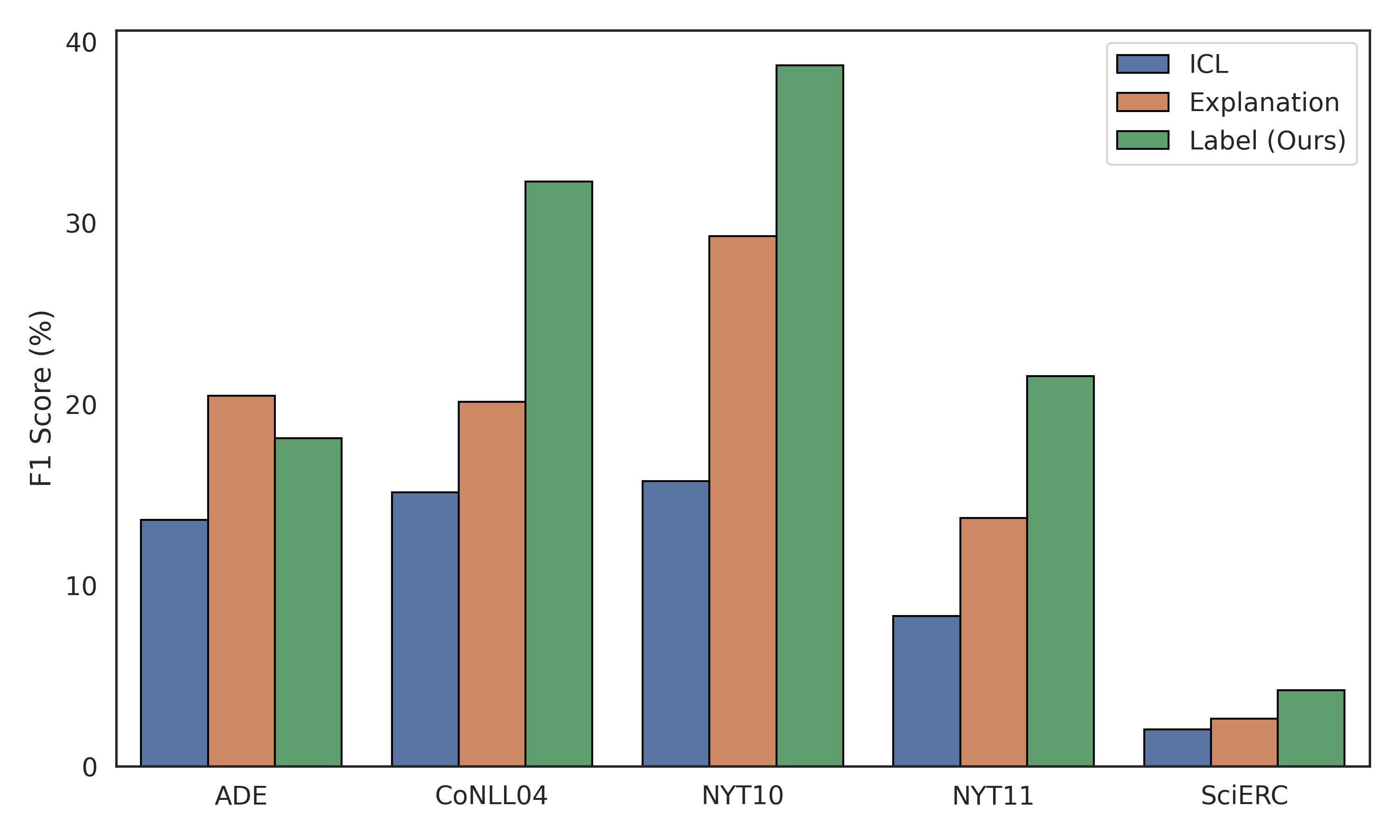}
    }
    \par\vspace{1mm}
    \subfloat[NER Task]{
        \centering
        \includegraphics[width=0.95\columnwidth]{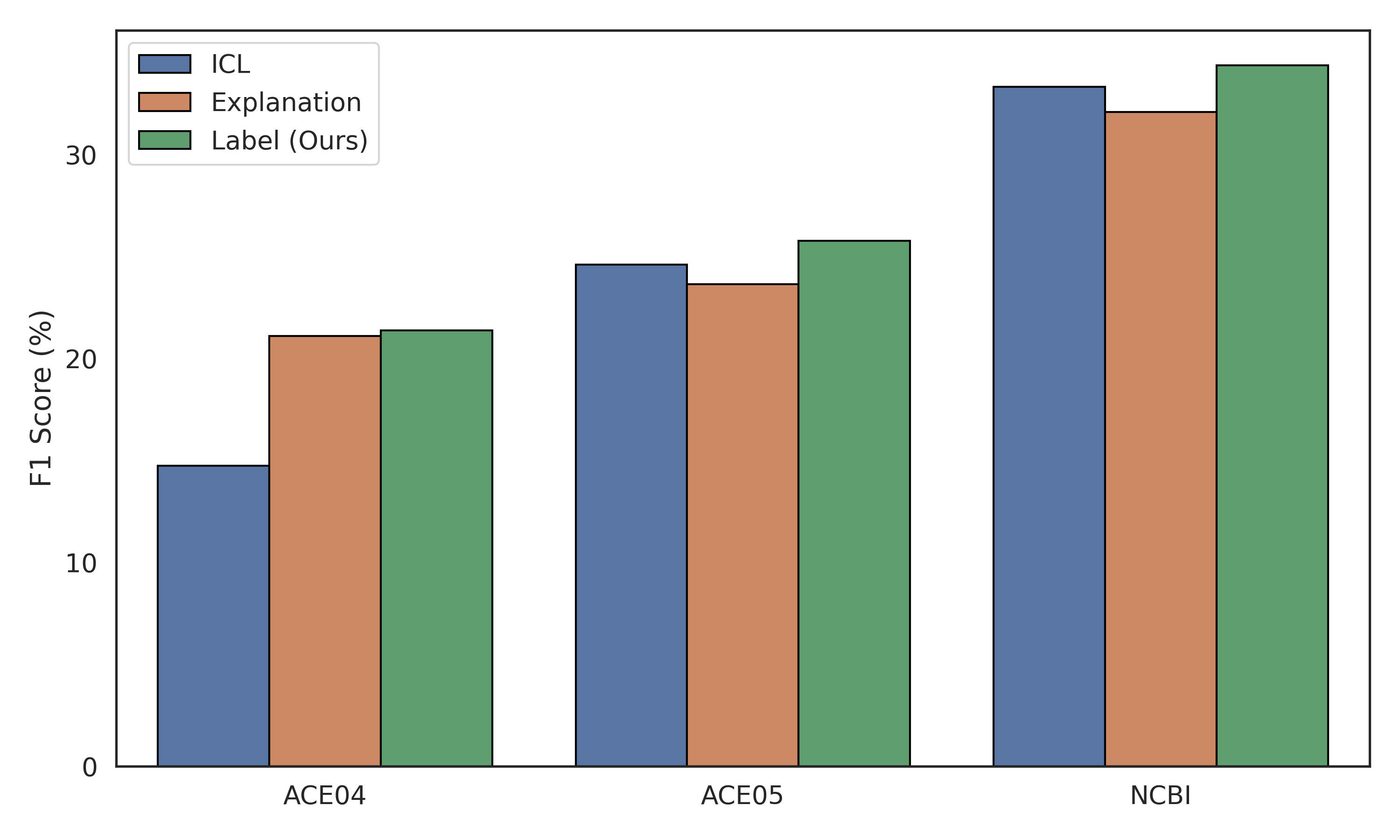}
    }
    \vspace{1mm}
    \caption{Comparative experimental results based on NER and RE tasks, evaluating the effectiveness of two negative sample identification methods: Label annotation and natural language explanation (Explanation). Here, Explanation refers to summarizing error reasons in the form of a natural language sentence, while Label refers to annotating error types in a structured label format (the method adopted in this paper).}

    \label{Fig.ablation_study} 
    \vspace{-5mm}
\end{center}
\end{figure}

To investigate the impact of structured features in mislabeled negative samples on experimental results, we conduct a comparative study between two types of negative samples: those annotated with structured labels (denoted as \textbf{Label}) and those annotated with a one-sentence explanation of the error reason generated by a large language model (denoted as \textbf{Explanation}). Experiments are conducted on both Named Entity Recognition (NER) and Relation Extraction (RE) datasets.

In our approach, we categorize and label model inference errors in detail, and reintroduce them as negative samples in In-Context Learning (ICL) demonstrations. This is intended to enhance the model's ability to recognize and correct error patterns. In contrast, the baseline \textbf{Explanation} method provides only a single-sentence summary of the error cause, without distinguishing between specific error types.

In most cases, both the \textbf{Explanation} and \textbf{Label} methods can enhance the information extraction capabilities of large language models. This can be attributed to the fact that negative samples provide certain error patterns, which help the model avoid similar mistakes and thereby reduce the overall error rate. However, on the ACE05 and NCBI datasets, the \textbf{ICL} method outperforms the Explanation approach. This may be because, for relatively simple NER tasks, the longer explanations generated by the Explanation method could increase the risk of hallucinations in large models, ultimately decreasing accuracy. Overall, the \textbf{Label} method consistently outperforms both the Explanation and ICL approaches, which strongly supports the effectiveness of our label design. In the following sections, we provide a detailed analysis of the Explanation and Label methods on each dataset.

\paragraph{Performance on NER Datasets}
Experimental results on three named entity recognition datasets---ACE04, ACE05, and NCBI---demonstrate that incorporating fine-grained label information consistently improves model inference performance to varying degrees. In the ACE04 dataset, the label-based approach yields a slight improvement (+0.28 percentage points). However, more substantial gains are observed on ACE05 and NCBI, with performance increases of +2.13 and +2.29 percentage points, respectively. These findings suggest that for NER tasks with complex structures or diverse entity types, detailed error annotations are more effective in helping large language models understand and correct their own inference biases. This effect is particularly pronounced in medical-domain datasets such as NCBI.

\paragraph{Performance on RE Datasets}
Except for the ADE dataset, all other RE datasets benefit significantly from the labeling approach, especially CoNLL04, NYT10, and NYT11, with F1 score improvements of +12.17, +9.42, and +7.56 percentage points, respectively. These results suggest that error types in relation extraction tasks are more diverse and complex, making it difficult for simple error summaries to effectively guide model improvements. In contrast, explicit error category labels help the model capture more fine-grained relational patterns, thereby significantly enhancing its error correction capability.

It is worth noting that on the ADE dataset, the labeling approach actually led to a performance drop of 2.62 percentage points. Upon deeper analysis, we find that the relation types in ADE are relatively homogeneous, primarily consisting of a single type: Drug-Adverse Effect. The task itself exhibits a simple structure, and the sources of errors are relatively concentrated. Under such circumstances, the multi-class error labels designed in the labeling method have limited utility. In fact, due to the insufficient number of error categories, the overall information content may even be reduced, making the method less effective than the \textit{Explanation} approach, which directly provides a concise summary of the error. 

The one-sentence explanation offered by the \textit{Explanation} method is able to guide the model's attention toward the core mistake in a direct manner under the simple scenario of ADE, significantly reducing redundant information and thereby improving inference accuracy.

This observation suggests that the effectiveness of error label design in relation extraction tasks is closely tied to the complexity of the task. For tasks with a single relation type and highly concentrated error patterns, concise and direct error explanations offer greater advantages. In contrast, for tasks involving diverse relation types and complex error patterns, fine-grained error labels are more beneficial in helping the model effectively reflect and optimize its reasoning process.

\subsection{Analysis of Label Definition Validity}

\begin{figure}[t]
\begin{center}
    \subfloat[RE Task]{
        \centering
        \includegraphics[width=0.95\columnwidth]{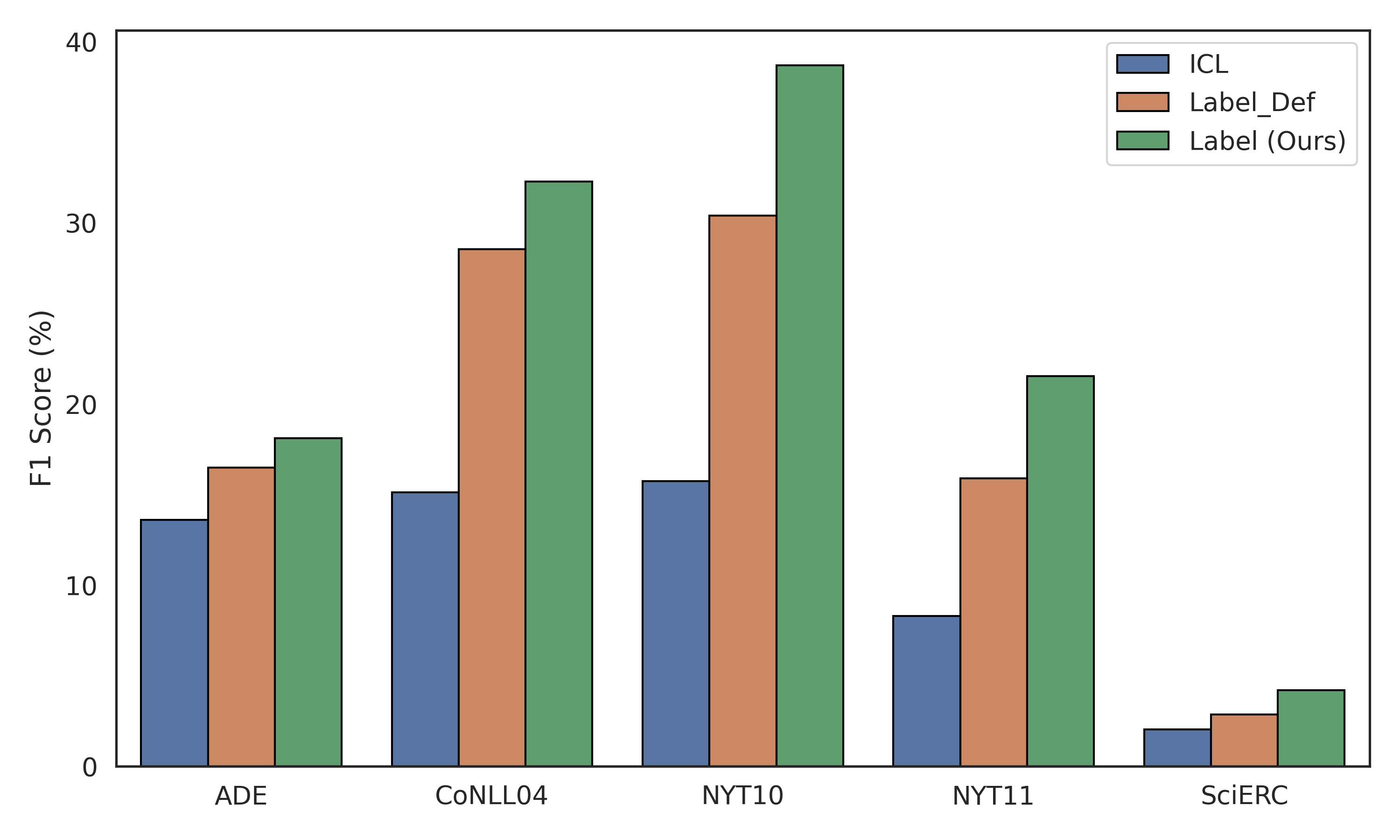}
    }
    \par\vspace{1mm}
    \subfloat[NER Task]{
        \centering
        \includegraphics[width=0.95\columnwidth]{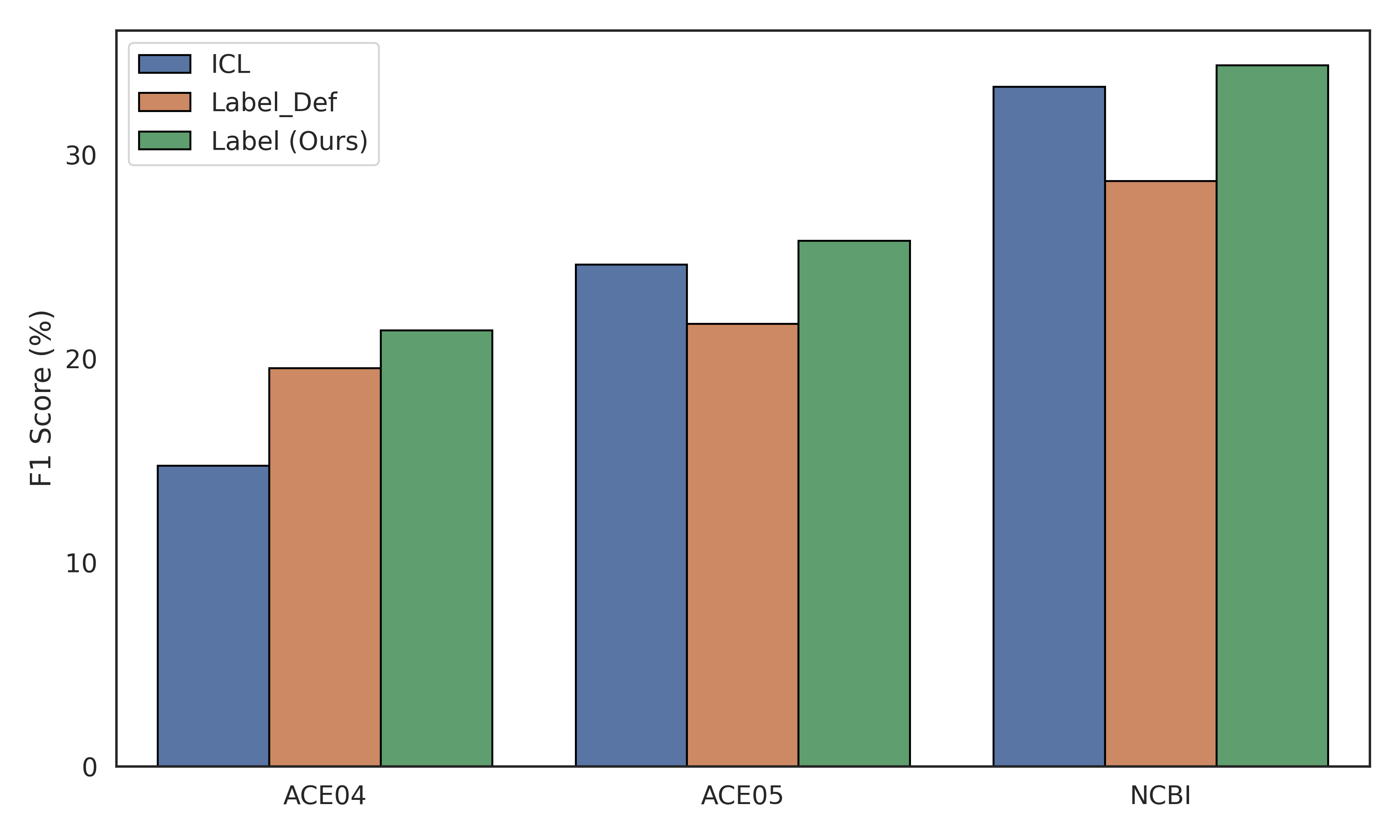}
    }
    \vspace{1mm}
    \caption{Comparison of experimental results under two settings: using only labels (Label) versus using labels accompanied by definitional explanations (Label\_Def). The results are evaluated on both NER and RE datasets. Here, Label\_Def refers to the setting where definitional explanations are provided for each label, while Label represents the setting where only the label itself is provided (which is the approach adopted in this paper).}

    \label{Fig.label_label_def} 
    \vspace{-5mm}
\end{center}
\end{figure}

To verify the impact of error-labeled in-context learning (ICL) examples on the inference capability of large language models, we designed two strategies for comparison: \textbf{Label} and \textbf{Label\_Def}. The \textbf{Label} strategy marks the specific type of error for each incorrect sample, while the \textbf{Label\_Def} strategy further provides a detailed explanation of the meaning of each label on top of the \textbf{Label} strategy.

Experiments were conducted on both Named Entity Recognition (NER) and Relation Extraction (RE) tasks, with F1 score used as the evaluation metric. The experimental results are summarized as follows.

Overall, both \textbf{Label} and \textbf{Label\_Def} strategies consistently improve the F1 scores of large language models on information extraction tasks across most datasets. This suggests that providing suitable negative samples enables large models to learn certain error patterns and avoid making the same mistakes. Detailed analysis for each dataset is presented below.

\paragraph{NER}

From the results of the NER task, the \textbf{Label} method consistently outperforms the \textbf{Label\_Def} method across all datasets, with an average F1 score improvement of over 4 percentage points, demonstrating a clear advantage. Upon analysis, this can be attributed to the fact that errors in entity boundary recognition and type classification in NER are highly categorizable. By directly using concise error labels, large language models are provided with clear correction directions. In contrast, the \textbf{Label\_Def} method introduces label definitions that may contain redundant information, potentially distracting the model from focusing on the core error categories, thereby weakening the effectiveness of inference enhancement.

Moreover, named entity recognition inherently emphasizes fine-grained boundary and type discrimination, where error types are relatively explicit. In this context, lengthy explanations offer limited benefits and may even lead the model to over-attend to irrelevant details, ultimately degrading performance.

\paragraph{RE}

In the relation extraction task, the Label method also achieves better performance across all datasets, with a particularly significant improvement in F1 score. This improvement is especially notable in complex relation scenarios such as NYT10 and NYT11, where the increase reaches +8.28 and +5.24 percentage points, respectively. This result indicates that the error patterns in relation extraction are complex and diverse. Directly providing structured error labels helps the model effectively align with error categories, enhancing the model's error correction and inference capabilities.

In contrast, the Label\_Def method, which introduces label explanations, did not bring the expected benefits in complex scenarios. Instead, it may have hindered the model's ability to focus on core error categories due to lengthy descriptions that distracted attention. Particularly in cross-domain, multi-relation datasets like the NYT series and SciERC, \textbf{concise and clear error labels are more effective in helping the model extract useful patterns than lengthy explanations}.

The performance on the ADE dataset remained relatively stable, with the Label method bringing only a slight improvement. Considering that the ADE relation type is singular (drug-adverse reaction), errors are more concentrated, and the marginal effect of label explanations is smaller, reflecting the close relationship between label design and task complexity.

\subsection{Validity of Negative Samples}

\begin{figure}[t]
\begin{center}
    \centering
    \includegraphics[width=1\columnwidth]{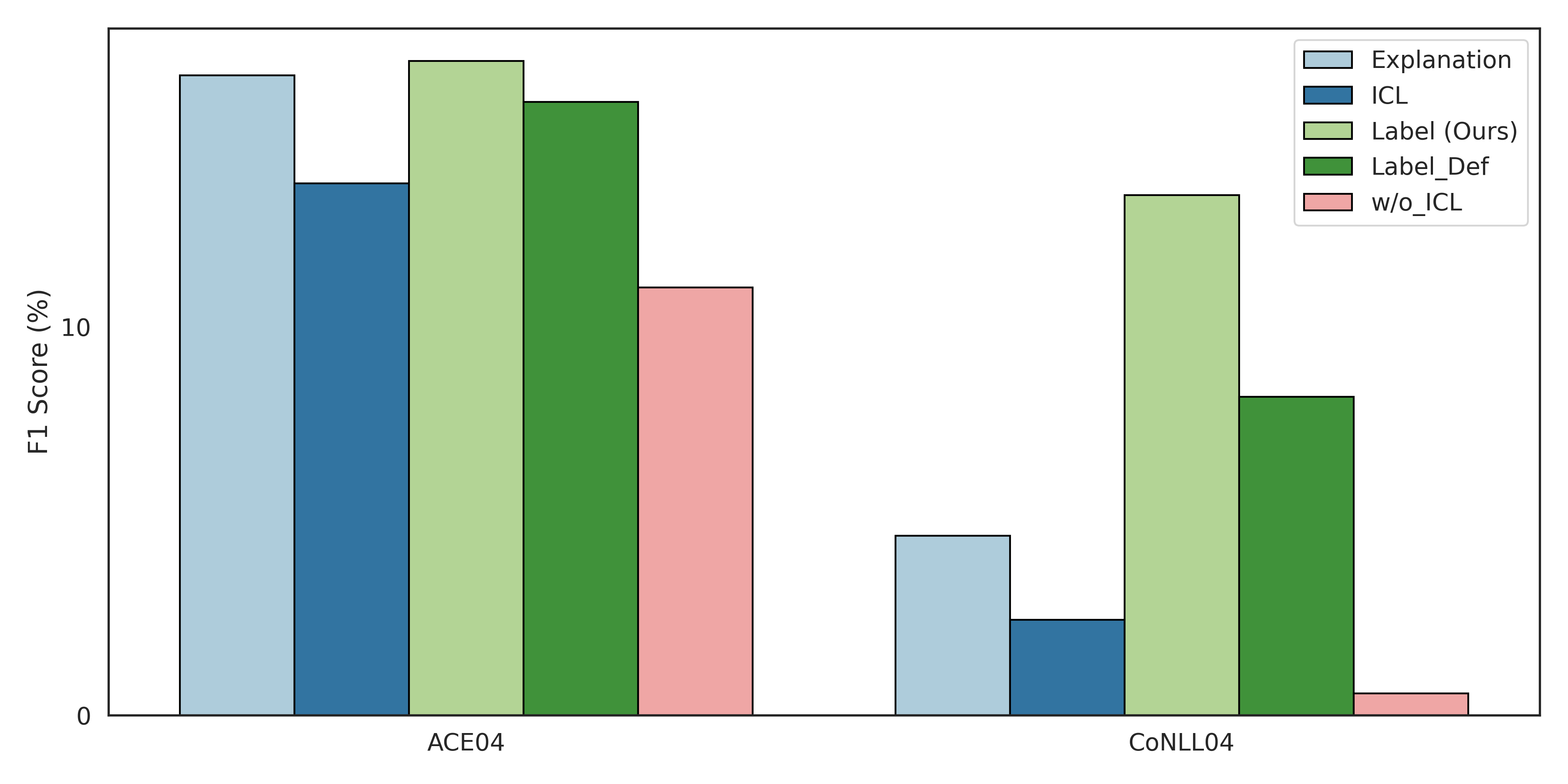}
    \vspace{-3mm}
    \caption{Comparison of RE and NER task performance of various methods under random retrieval strategy. We selected the ACE04 and CoNLL04 datasets to evaluate the performance of different methods. w/o\_ICL indicates that ICL is not used, and the large model performs zero-shot inference directly on the test text.}

    \label{Fig.random_ACE04_CoNLL04} 
    \vspace{-5mm}
\end{center}
\end{figure}

We employed a random retrieval strategy on the ACE04 and CoNLL04 datasets to observe the comparison between our method and others under random retrieval. 

To further validate the robustness and generalization ability of our proposed LC-ICL method, we designed a random retrieval (Random Retrieval) scenario, where we randomly select examples from the training set as support samples to replace similarity-based retrieval methods. This was done to evaluate the performance differences of various methods without effective retrieval support.

As shown in the figure, under the random retrieval strategy, our method still demonstrates significant advantages. In particular, the label-based method outperforms other baseline methods on both the NER and RE tasks:

In the ACE04 (NER) task, the F1 score of the Label method reached 16.84\%, significantly higher than Explanation (16.47\%), ICL (13.69\%), and w/o\_ICL (11.01\%).
In the CoNLL04 (RE) task, the Label method performed especially well, with an F1 score of 13.39\%, far surpassing Explanation (4.62\%), ICL (2.45\%), and w/o\_ICL (0.56\%).
Moreover, \texttt{Label\_Def} (label with definition) showed a slight decline compared to the pure Label method, indicating that under random retrieval, adding too much explanatory information may weaken the effectiveness of the contrastive signal.

From the random retrieval experiment, it can be observed that the label-based contrastive example construction strategy we proposed demonstrates good robustness. Even in the absence of semantic retrieval support and with poor-quality supporting sample information, the Label method still significantly improves the model's reasoning and discrimination abilities. This phenomenon further validates the core advantage of our method, which lies in explicit error signal labeling and structured supervision, rather than relying on the retrieved "good examples."

In summary, the random retrieval experiment fully verifies the effectiveness and generalization of the \ourmethod{} method, indicating that reasonably designed label information is more effective in helping large models understand task objectives and improve performance than solely relying on sample similarity.

\subsection{Case study}

\begin{figure}[ht]
\begin{center}

    \centering
    \includegraphics[width=1\columnwidth]{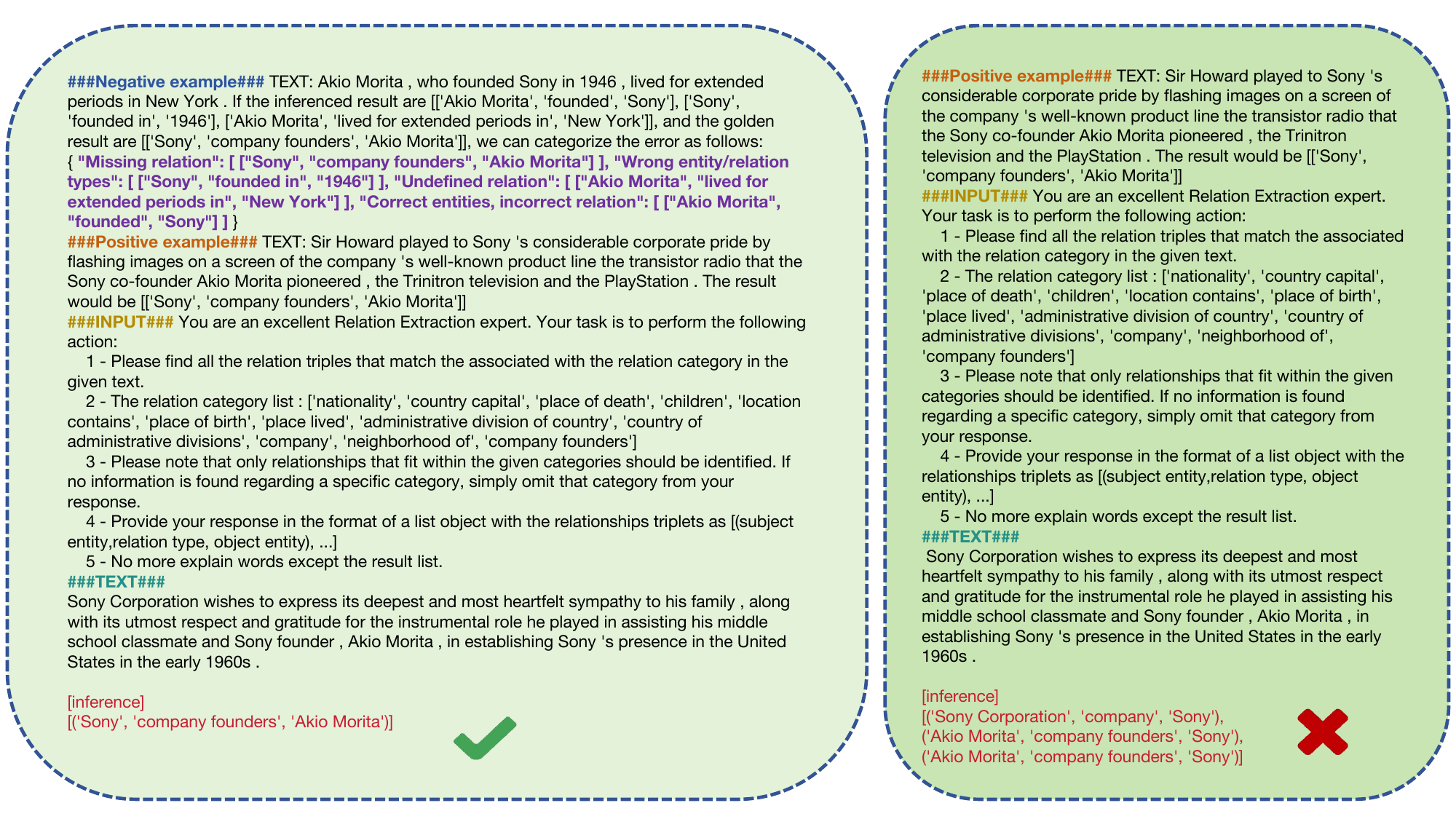}
    \vspace{-3mm}
    
    \caption{A comparison of prompts between \ourmethod{} and ICL methods on the same RE task. The traditional ICL approach (right) includes positive examples, instructions, and a test instance. In contrast, \ourmethod{} (left) augments this setup by additionally incorporating a negative example, which not only presents an instance where the large language model makes a reasoning error, but also provides a set of labels indicating the reasons behind the error. For the NER task and additional examples, please refer to the appendix.}

    \label{Fig.example1}   
    \vspace{-5mm}

\end{center}
\end{figure}
We selected a number of representative test cases to further demonstrate the effectiveness of our proposed method, as shown in Figure~\ref{Fig.example1}. 

Taking Example 1 as an illustration, we present a contrastive set of positive and negative examples in the relation extraction (RE) task that are more similar to the test instance. 

In this case, using only the ICL method (right side of the figure) indeed enables the model to effectively identify the entities "Sony" and "Akio Morita", but it fails to accurately recognize their relationship. In contrast, our method (left side of the figure) allows the large language model to learn that the reasoning outcome \["Akio Morita", "founded", "Sony"\] corresponds to the label "Correct entities, incorrect relation" after being exposed to negative samples. As a result, the model becomes more attentive to the correctness of the relation. 

By learning correct patterns from positive examples and simultaneously learning to avoid erroneous patterns from negative examples, the large model is better able to derive accurate conclusions.

In addition, we also discuss typical cases in the named entity recognition (NER) task; more examples can be found in Appendix~\ref{sec:appendix_c}.

\section{Related Work}
\paragraph{Generative Information Extraction}
In recent years, a large number of representative supervised modeling methods have been proposed for sub-tasks in information extraction, such as named entity recognition (NER) and relation extraction (RE). Early works primarily focused on understanding-based architectures, such as the LSTM-CRF model~\cite{Lample2016} and RE methods based on dependency tree modeling~\cite{tang2018}, which effectively capture both local and syntactic features.

\cite{9521975} leveraged the mutual influence between RE and NER tasks to design a dynamic cross-task model that further enhances the model's information extraction capabilities. However, supervised modeling methods typically require fine-tuning on specific datasets, resulting in limited generalization ability. 

Meanwhile, the rise of generative Transformer architectures has introduced a new paradigm for information extraction, where the task is reformulated as a text generation problem. Works such as the Crop model~\cite{crop}, the MCL-NER framework~\cite{mcl-ner-moying-2023}, and zero-shot IE methods based on ChatGPT~\cite{zeroshot_IE_ChatGPT} have all demonstrated the strong potential of large models in extraction tasks.

Overall, although generative large language models exhibit strong potential and diverse developmental trends in information extraction tasks, their performance remains suboptimal when directly applied to extraction tasks without fine-tuning.

\vspace{-1mm}
\paragraph{In-context Learning} 
In recent years, in-context learning (ICL) has garnered extensive attention and application in large-scale language models. By constructing high-quality exemplars, ICL significantly enhances the models' reasoning and task execution capabilities~\cite{learn_to_select,retrive-ICL-2022,ICL-survey-dong2023}. To further improve the reasoning abilities of language models, existing studies have shown that relying solely on the models' inherent reasoning mechanisms is insufficient to fully unleash their potential. Instead, constructing detailed chains of thought can notably boost the performance of reasoning models~\cite{ge2025innatereasoningenoughincontext}, underscoring the critical role of context construction in influencing model performance. In addition, methods such as nearest-neighbor retrieval~\cite{KATE-knn-2022}, task-aware retrieval strategies, and label-guided reasoning representations~\cite{GPT-RE-23} have further optimized the process of exemplar selection.

Recently, substantial progress has also been made in instruction-following tasks. \cite{lin2023unlocking} proposed the URIAL method, which achieves effective alignment for base language models using only three stylized in-context examples. Although URIAL demonstrates notable improvements in instruction-following capabilities, \cite{zhao2024context} pointed out that it still underperforms fine-tuned models in multi-turn interactive scenarios.

Overall, in-context learning exhibits great potential in enhancing the reasoning ability and task performance of large language models. However, its limitations have also gradually emerged, primarily due to the lack of negative feedback mechanisms, which restricts further performance improvement.

\vspace{-1mm}

\paragraph{Hard negative sample}
Hard negative samples have a long-standing history of applications in machine learning. In the early days, \cite{mikolov2013distributed} introduced the Word2Vec model, which employed negative sampling to train word embeddings, enabling more efficient learning of textual features. Similarly, \cite{devlin2019bert} enabled the model to predict masked tokens while treating other tokens as negative samples, significantly improving performance on a range of NLP tasks such as question answering, sentiment analysis, and text classification. A current research focus is whether erroneous information generated by large language models (LLMs) during inference can be systematically categorized and utilized as negative samples to enhance model capabilities. Relevant scholars have conducted systematic analyses and discussions on this topic. For instance, \cite{simhi2024distinguishing} presented a detailed taxonomy of hallucinations in LLMs, distinguishing between "ignorance-based errors" caused by knowledge gaps and "knowingly wrong" errors where the model possesses the correct knowledge yet generates incorrect outputs. This distinction provides a crucial theoretical foundation for error detection and intervention. In addition, \cite{lu2023error} proposed a prompting method that integrates chain-of-thought reasoning with error analysis. By simulating human evaluation systems to conduct fine-grained error categorization in translation outputs, the method not only improves evaluation performance but also enhances model interpretability. In the fields of Named Entity Recognition (NER) and Relation Extraction (RE), \cite{10856568,10488145} skillfully applied similar principles, leveraging comparison-based learning to further boost accuracy in information extraction tasks. Furthermore, \cite{mo2024c-icl} proposed a theoretical framework based on consistency learning to obtain high-quality negative samples, thereby advancing the information extraction capabilities of LLMs.

Motivated by these observations, the core goal of this study is to exploit specific error patterns in LLM-generated content as learning signals. These patterns are fed back to the model in the form of hard negative samples, with the aim of improving the model's performance on targeted tasks.

\vspace{-1mm}

\section{Conclusion}
In this work, we introduce \ourmethod{}, contrastive in-context learning for few-shot information extraction, including right/positive and wrong/negative demonstrations. 
In addition through type instruction demonstrations prompt mention tags in the IE task. 
From the contrastive samples, the LLMs could obtain effective information and indirect but positive, valuable additional knowledge for IE tasks. 
Besides, our method adopts semantic similarity retrieval strategies to retrieve in-context examples better suited for the current sentence and task, significantly improving IE performance.
Extensive experiments prove the effectiveness of \ourmethod{} on various benchmarks.
\section*{Limitations}
Despite its contributions, this study has the following limitations: (1) It primarily investigates the contextual learning capabilities in few-shot Named Entity Recognition (NER) and Relation Extraction (RE) tasks, while the applicability of this paradigm to other Information Extraction (IE) tasks remains underexplored; (2) We employed a commonly used sentence embedding similarity approach to retrieve samples, yet there may exist other, more diverse strategies for selecting appropriate positive and negative samples; (3) Our evaluation was conducted using English datasets on large models trained in English, and further exploration on datasets or models in other languages, such as Chinese, was not carried out.

\bibliography{custom}
\bstctlcite{IEEEexample:BSTcontrol}

\clearpage
\appendix
\section{Implementation Experiment}
\label{sec:appendix_a}
\subsection{Dataset Statistics}
To facilitate a thorough evaluation, we incorporate a diverse collection of datasets spanning both NER and RE tasks, comprising three widely-used NER benchmarks and five representative RE benchmarks.
The detailed statistics of these datasets , including the number of entity and relation types, along with the instance distributions across the training, development, and test sets , are presented in Table \ref{tab:datasets_anlasis}.
This comprehensive overview not only illustrates the characteristics of each dataset but also underscores the robustness and generalizability of our evaluation framework.
\begin{table}[ht]
\caption{Statistics of NER and RE Datasets.}
\label{tab:datasets_anlasis}
\begin{adjustbox}{width=1\columnwidth,center}
\begin{tabular}{ccccccc}
\toprule
\multicolumn{2}{c}{Datasets}   & \tabincell{c}{Entity \\ Types}  & \tabincell{c}{Relation \\ Types} & Train & Dev  & Test \\ \hline
\multirow{3}{*}{NER} 
                     & ACE04   & 7            & /              & 6202  & 745  & 812  \\
                     & ACE05   & 7            & /              & 7299  & 971  & 1060 \\
                     & NCBI & 1            & /              & 5433 & 924 & 941 \\ \hline
\multirow{5}{*}{RE}  & CoNLL04 & 4            & 5              & 922   & 231  & 288  \\
                     & NYT10     & 3            & 24             & 56196 & 5000 & 5000 \\
                     & NYT11     & /           & 12             & 62648 & 149 & 369 \\
                     & ADE   & /            & 1              & 3417 & 427 & 428 \\

                     & SciERC  & 6            & 7              & 1861  & 275  & 551 \\\bottomrule
\end{tabular}
\end{adjustbox}
\end{table}

\subsection{Implementation Experiment Details}
All experiments are conducted using the PyTorch deep learning framework on NVIDIA Tesla A100 GPUs.
To ensure optimal model performance, we carefully select the experimental settings and hyperparameters.
These include the maximum sequence length, number of beams for beam search, as well as the top-p and temperature values that control the randomness during generation.
The parameters are detailed in Table \ref{tab:parameters}.
\begin{table}[ht]
\caption{The main parameters of our method \ourmethod{} based on Llama.}
\label{tab:parameters}
\begin{adjustbox}{width=0.8\columnwidth,center}
\begin{tabular}{lc}
\toprule
Parameters                        & Values   \\ \hline
Max Sequence Length               & 8192      \\
Num\_beams                        & 1         \\
Do\_sample                        & True      \\
Top\_p                            & 0.85       \\  
Temperature                       & 0.3      \\ \bottomrule
\end{tabular}
\end{adjustbox}
\end{table}
\section{Supplementary Label Definition Explanation}
\label{sec:appendix_b}
\begin{figure*}[ht]
\begin{center}
    \centering
    \includegraphics[width=2\columnwidth]{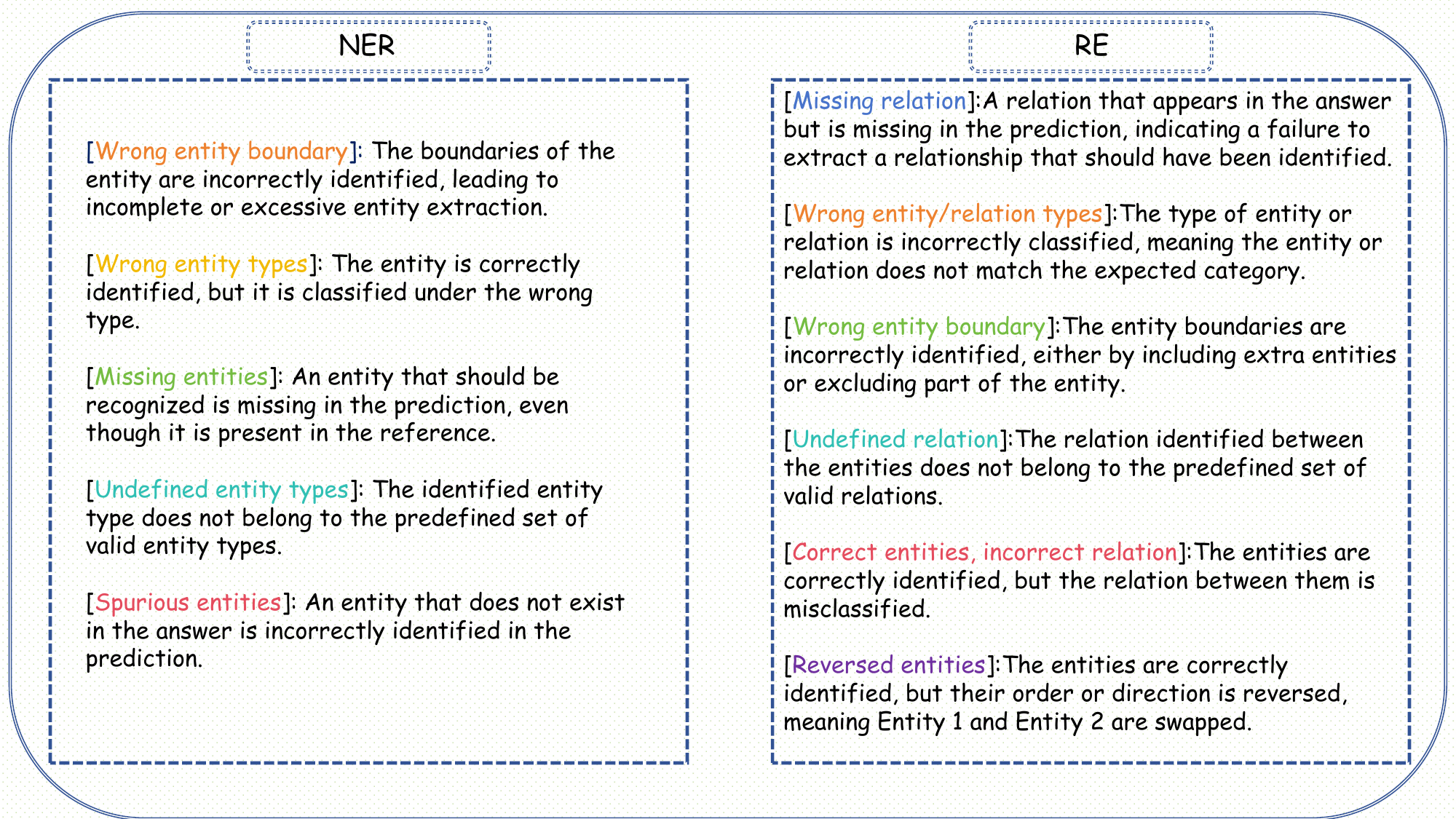}
 
    \vspace{1mm}
    \caption{Label Categorization Illustration.
In the Relation Extraction (RE) task, labels are categorized into six classes, while in the Named Entity Recognition (NER) task, they are divided into five classes, each accompanied by corresponding label definitions.}
    \label{Fig.label_def} 
    \vspace{-5mm}
\end{center}
\end{figure*}

Figure \ref{Fig.label_def} illustrates  all label types and their corresponding definitions used for negative samples in our method.
For the Named Entity Recognition (NER) task, we categorize common sources of errors into five label types: \textit{Wrong entity boundary}, \textit{Wrong entity types}, \textit{Missing entities}, \textit{Undefined entity types}, and \textit{Spurious entities}.
For the Relation Extraction (RE) task, we classify common error sources into six label types: \textit{Missing relation}, \textit{Wrong entity/relation types}, \textit{Wrong entity boundary}, \textit{Undefined relation}, \textit{Correct entities, incorrect relation}, and \textit{Reversed entities}.
Each label type is immediately followed by an explanation of its definition.
When employing the Label-Def method, as shown in experiment \ref{Fig.label_label_def}, we provide the model with all label types and their definitions as prompts, depending on whether the current task is RE or NER.

\section{Supplementary Case Study}
\label{sec:appendix_c}
\begin{figure*}[ht]
\begin{center}
    \subfloat[Example 2]{
        \includegraphics[width=0.95\textwidth]{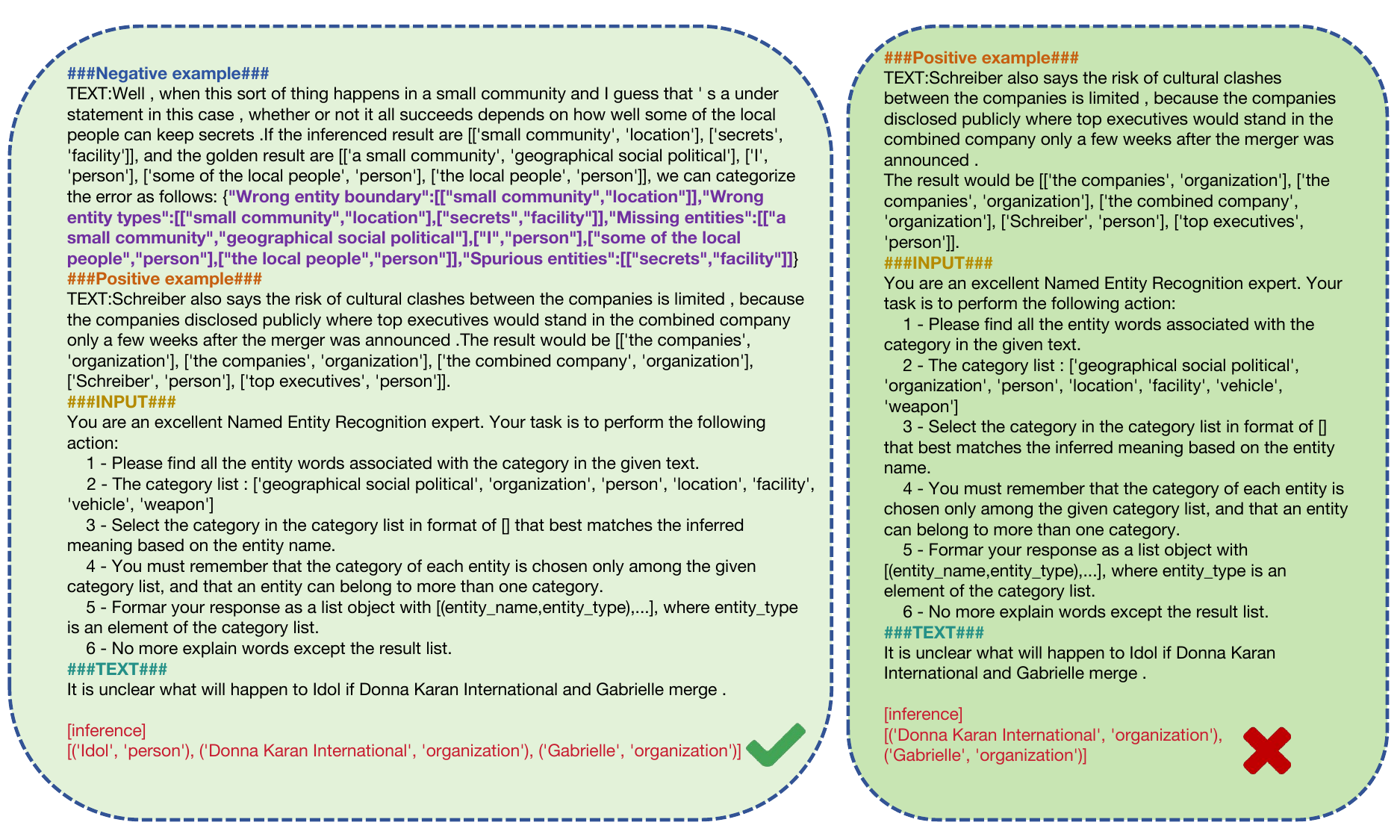}
        \label{Fig.example2}
    }
    \par\vspace{2mm}
    \subfloat[Example 3]{
        \includegraphics[width=0.95\textwidth]{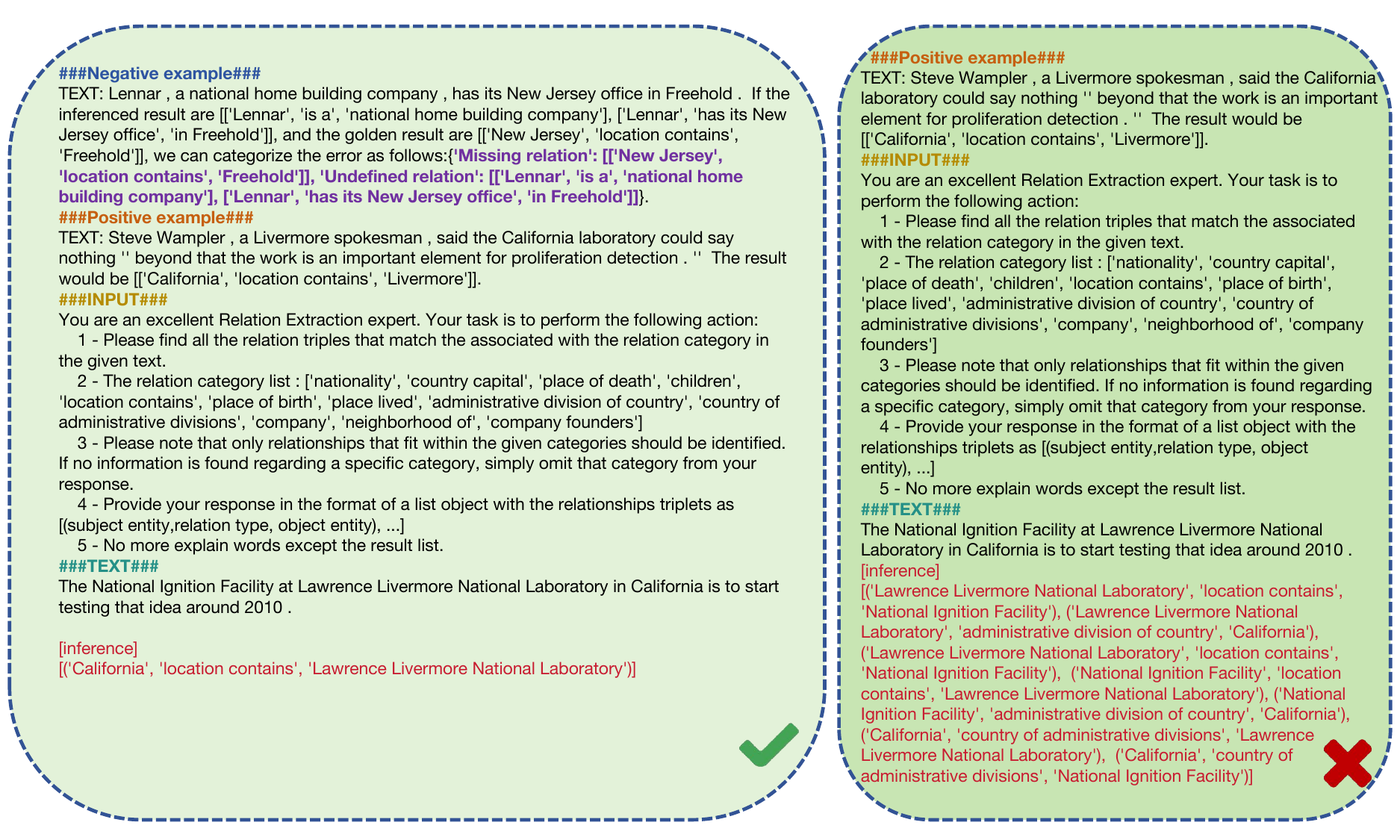} 
        \label{Fig.example3}
    }
    \caption{A supplementary case study on label-guided contrastive in-context learning is presented. The left side shows the results of our method, while the right side presents those of the standard ICL approach. The purple text highlights the annotated reasons for errors in negative samples. Figure \ref{Fig.example2} illustrates the results of the Named Entity Recognition (NER) task, and Figure \ref{Fig.example3} shows the results of the Relation Extraction (RE) task. }
    \label{Fig.other_case_study}
    \vspace{-15pt}
\end{center}
\end{figure*}

In this section, we present additional examples from the NER and RE tasks, as shown in Figure \ref{Fig.other_case_study}.
For each example, the left side displays the results using the LC-ICL method with injected error labels, while the right side shows the results from the traditional ICL method.

In Example 2, when using only the traditional ICL method, the model fails to capture the relation ("Idle", "person").
In contrast, our method provides negative samples that similarly omit three "person" entities---["I", "person"], ["some of the local people", "person"], and ["the local people", "person"]---and labels them as \textit{Missing entities}.
By learning from these negative samples, the model is able to correctly recognize and avoid similar cases of entity omission, ultimately producing the correct output.

In Example 3, the test sample is a relatively challenging sentence featuring three complex location entities: "The National Ignition Facility," "Lawrence Livermore National Laboratory," and "California."
Using the traditional ICL method, the model learns some basic patterns for location relations; however, in the presence of complex location structures, it becomes more prone to hallucinations, generating a large number of incorrect relations.
By incorporating negative samples, our method provides the model with error label information indicating missing correct relations as well as distracting location relations.
This additional supervision enables the model to better avoid hallucinations and significantly improves prediction accuracy.

\end{document}